\documentclass[conference]{IEEEtran}

\usepackage{cite}
\usepackage{amsmath,amssymb,amsfonts}
\usepackage{algorithm}
\usepackage{algorithmic}
\usepackage{graphicx}
\usepackage{textcomp}
\usepackage{xcolor}
\usepackage{booktabs}
\usepackage{multirow}
\usepackage[caption=false,font=footnotesize]{subfig}
\usepackage{url}
\usepackage[hidelinks]{hyperref}


\def\BibTeX{{\rm B\kern-.05em{\sc i\kern-.025em b}\kern-.08em
    T\kern-.1667em\lower.7ex\hbox{E}\kern-.125emX}}

\begin{document}

\title{BlockServe: Block-Grained Continuous Batching for High-Throughput Diffusion LLM Serving}

\author{\IEEEauthorblockN{Yuanjie Zhu, Liangwei Yang, Ke Xu, Weizhi Zhang, Shanghao Li, Zihe Song, and Philip S. Yu}
\IEEEauthorblockA{\textit{University of Illinois Chicago, USA}\\
\{yzhu224, lyang84, kxu25, wzhan42, sli261, zsong29, psyu\}@uic.edu}}

\maketitle

\begin{abstract}
Efficient serving of diffusion large language models (dLLMs) is hindered by convergence heterogeneity: when batching multiple requests, different sequences converge at different rates, causing faster requests to stall behind slower stragglers and introducing compute bubbles and tail latency. We present BlockServe, a continuous batching framework that integrates block-grained scheduling---immediately evicting completed requests at block boundaries---with mixed-state execution that extends dual cache and parallel decoding to heterogeneous batches via gather-scatter indexing. Furthermore, a compute-aware admission controller expands effective batch capacity through token-budgeted refill. On Dream and LLaDA across five benchmarks, BlockServe achieves 1.9--10.6$\times$ throughput over Fast-dLLM with comparable generation quality, establishing block-grained scheduling as a foundation for high-throughput offline dLLM inference.

\end{abstract}

\begin{IEEEkeywords}
diffusion large language models, continuous batching, model serving, block-grained scheduling, throughput
\end{IEEEkeywords}

\section{Introduction}
\label{sec:intro}

The deployment of Large Language Models (LLMs) has shifted focus from model architecture to efficient serving systems~\cite{holmes2024deepspeed}.
In the autoregressive (AR) paradigm, continuous batching~\cite{yu2022orca,kwon2023efficient} has become the industry standard, scheduling at the granularity of a single token to maintain high GPU utilization.
However, the emergence of Diffusion LLMs (dLLMs)~\cite{nie2025large,ye2025dream} introduces a fundamental paradigm shift. Unlike AR models that generate tokens sequentially, dLLMs employ a parallel denoising process, generating entire blocks of tokens iteratively. While recent works like Fast-dLLM~\cite{wu2025fast} have successfully optimized dLLMs for low-latency single-request inference, efficient batched serving for these models remains challenging when requests converge at different rates across block boundaries.

We term this phenomenon convergence heterogeneity. In autoregressive serving, output length variation is the primary source of scheduling heterogeneity, which token-level continuous batching handles by preempting at every generation step~\cite{agrawal2023sarathi,agrawal2024taming}. dLLMs, however, generate in discrete blocks, and requests reach completion at different block boundaries even under the same generation budget. As a result, some requests can finish early while others continue denoising, creating mixed completion states within a single batched serving iteration. In current dLLM batching, the entire batch is gated by the slowest request (the ``straggler''), forcing already-completed requests to remain idle while occupying resources in the active batch. The resulting ``compute bubbles''---idle time where GPU resources are underutilized---lead to severe long-tail latency, calling for a dedicated block-level batch scheduling framework.

To address these challenges, we propose BlockServe, a specialized serving framework that enables high-throughput continuous batching for dLLMs without altering the underlying model architecture. BlockServe integrates a block-grained scheduler with a mixed-state memory manager, preventing stragglers from stalling the batch while enabling the execution of requests at different block indices within a unified dense tensor. A compute-aware admission controller under a token budget further adapts concurrency to workload geometry. These components form a block-centric execution loop: completed requests are reclaimed at block boundaries, heterogeneous in-flight requests remain executable in a shared dense batch, and the recovered capacity is then converted into additional admissions. Our contributions are summarized as follows:
\begin{itemize}
    \item We introduce a block-grained scheduler that operates at block granularity, immediately evicting completed requests via block-wise completion checks to reduce straggler-induced compute bubbles in batched diffusion inference.
    \item We design a mixed-state memory manager that materializes unified dense tensors with explicit position alignment, extending dual cache and parallel decoding to heterogeneous batches without requiring custom kernels.
    \item We propose a compute-aware admission controller that replaces fixed batch sizes with a token budget, dynamically adapting concurrency to workload geometry and enabling a 2--4$\times$ increase in effective batch capacity.
    \item On Dream and LLaDA across five benchmarks, BlockServe achieves 1.9--10.6$\times$ throughput over Fast-dLLM with comparable generation quality.
\end{itemize}

\section{Method}
\label{sec:method}

\subsection{Problem formulation}
\label{sec:method:formulation}

We consider the concurrent serving of a diffusion LLM processing a batch of user requests.
Unlike autoregressive generation, dLLMs produce the sequence via an iterative denoising process.
For a specific request $r_i$ with prompt length $p_i$ and target generation length $g_i$, the generation workload is decomposed into $K_i = \lceil g_i / L \rceil$ fixed-length execution blocks of length $L$ along the generated sequence.
Each block is iteratively denoised over $S$ diffusion steps.
The state of request $r_i$ is characterized by its current block index $b_i \in \{0, \dots, K_i-1\}$.
The core objective of BlockServe is to maximize system throughput by maintaining an active set $\mathcal{A}$ of mixed-state requests---where $r_i, r_j \in \mathcal{A}$ may be at distinct block indices $b_i \neq b_j$---while respecting hardware memory constraints.

\subsection{System overview}
\label{sec:method:overview}

BlockServe operates as a continuous iteration loop centered on block-grained scheduling and mixed-state execution.
In each iteration, the Block-Grained Scheduler (Section~\ref{sec:method:scheduler}) advances all active requests by exactly one block-cycle, regardless of their individual lifecycle stages, while the Mixed-State Memory Manager (Section~\ref{sec:method:memory}) maps these heterogeneous requests into a unified dense tensor layout for vectorized execution.
To further expand effective batch capacity, the Compute-Aware Admission Controller (Section~\ref{sec:method:admission}) selects candidates from the pending queue $\mathcal{P}$ to populate the active set $\mathcal{A}$ under a length-aware token budget.

\begin{figure*}[t]
    \centering
    \includegraphics[width=\textwidth]{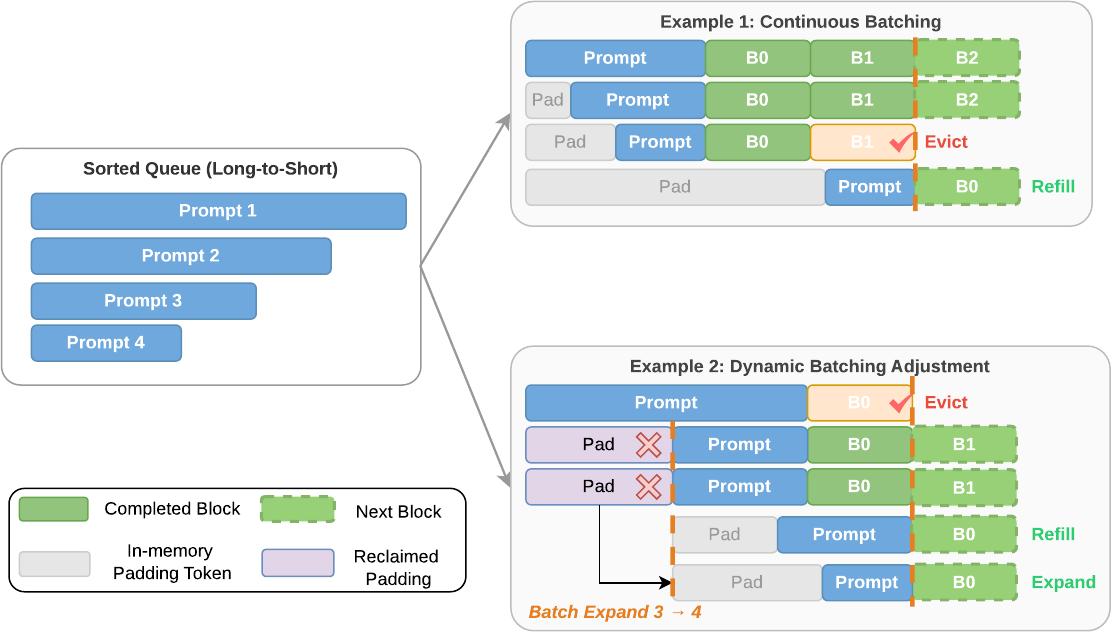}
    \caption{Overview of BlockServe's continuous batching framework. Requests from a length-sorted queue (left) are batched at block granularity; dashed-green cells indicate the next block to be processed. Example~1 (top): eviction and refill upon request completion. Example~2 (bottom): dynamic batch expansion via bounding-box shrinkage.}
    \label{fig:overview}
\end{figure*}

Figure~\ref{fig:overview} illustrates how these stages interact. In Example~1, when request 3 finishes its last block, the scheduler evicts it at the block boundary and the admission controller refills the vacant slot with request 4 from the sorted queue. Example~2 demonstrates how the batch can expand: after request 1 completes and is evicted, its large row width no longer constrains the bounding box, so previously required padding (purple) is reclaimed, freeing budget for two additional admissions and expanding the batch from three to four.

\subsection{Block-grained scheduling}
\label{sec:method:scheduler}

\begin{algorithm}[t]
\caption{BlockServe scheduling loop}
\label{alg:scheduling}
\begin{algorithmic}[1]
\STATE \textbf{Input:} Pending queue $\mathcal{P}$ (sorted by $p_i$) with $(K_i, b_i) = (\lceil g_i / L \rceil, 0)$ for each request $r_i$; Active set $\mathcal{A} \leftarrow \emptyset$; block length $L$; token budget $\beta_{\text{budget}}$
\WHILE{$\mathcal{P} \neq \emptyset$ \textbf{or} $\mathcal{A} \neq \emptyset$}
    \STATE Admit requests from $\mathcal{P}$ into $\mathcal{A}$ while $\text{Cost}(\mathcal{A} \cup \{r_{\text{cand}}\}) \le \beta_{\text{budget}}$ \COMMENT{\S\ref{sec:method:admission}}
    \STATE $\mathbf{Y} \leftarrow \text{ExecBlockCycle}(\mathcal{A})$ \COMMENT{\S\ref{sec:method:memory}}
    \FOR{each request $r_i \in \mathcal{A}$}
        \STATE Update $r_i$ output with $\mathbf{Y}[i]$ and set $b_i \leftarrow b_i + 1$
    \ENDFOR
    \STATE Remove from $\mathcal{A}$ any request with $b_i \ge K_i$ or $\text{CheckEOS}(r_i)$
\ENDWHILE
\end{algorithmic}
\end{algorithm}

BlockServe's core innovation is the shift from request-level locking to block-level preemption.
Each scheduling quantum corresponds to one block-cycle of $S$ denoising steps.
Algorithm~\ref{alg:scheduling} outlines the protocol. In each iteration, the admission controller populates the active set $\mathcal{A}$ from the pending queue $\mathcal{P}$ under the token budget constraint (Section~\ref{sec:method:admission}). The scheduler then executes one block-cycle for the batch in parallel (Section~\ref{sec:method:memory}); this execution is block-agnostic, as requests at different generation stages are advanced within the same block-cycle. After execution, the scheduler performs block-boundary completion checks: any request that has reached its block limit $K_i$ or produced an end-of-sequence token is immediately evicted from $\mathcal{A}$. This fine-grained preemption mitigates the straggler effect by reclaiming resources at each block boundary rather than waiting for the entire batch to synchronize.

\subsection{Compute-aware admission control}
\label{sec:method:admission}

To maximize hardware utilization without causing Out-Of-Memory (OOM) errors, we abandon fixed batch sizes in favor of a token budget policy. Since standard GPU kernels operate on regular dense tensors, serving unequal-length requests within the same batch requires padding, which leads to memory waste.
We define the memory cost of an active batch $\mathcal{A}$ based on the bounding box of the dense tensor that must be materialized at each iteration. Since BlockServe only allocates space for prompt tokens and blocks that have been completed or are currently being denoised, the effective row width of request $r_i$ at block index $b_i$ is $p_i + (b_i + 1) \times L$. The batch cost is:
\begin{equation}
\label{eq:cost}
    \text{Cost}(\mathcal{A}) = |\mathcal{A}| \times \max_{r_i \in \mathcal{A}}\!\bigl(p_i + (b_i + 1) \times L\bigr)
\end{equation}
Prompt length and block progress determine the bounding box. When the request defining the current maximum completes and is evicted, the bounding box shrinks---freeing budget for new admissions without explicit memory reclamation.
Minimizing padding waste requires reducing the gap between each request's row width and the batch-wide maximum.

To approximate this, we implement a length-aware greedy batching strategy.
The pending queue $\mathcal{P}$ is maintained in sorted order by prompt length.
During the refill phase, the scheduler searches for candidates $r_{\text{cand}}$ whose initial row width $p_{\text{cand}} + L$ (starting at block~0) fits within the current bounding box without expanding it (``sand filling'').
A candidate is admitted only if $\text{Cost}(\mathcal{A} \cup \{r_{\text{cand}}\}) \le \beta_{\text{budget}}$, where $\beta_{\text{budget}}$ is determined by offline profiling of peak GPU memory utilization.
This enables BlockServe to safely surpass static batch size limits, adapting concurrency to workload geometry.

\subsection{Mixed-state memory management}
\label{sec:method:memory}

Efficient mixed-state execution requires reconciling request raggedness with the physical density required by GPU kernels.
BlockServe achieves this through two coordinated mechanisms: Logical Positional Alignment and Vectorized Mixed-Schedule Execution.

\paragraph{Logical positional alignment}
We materialize the bounding box (Eq.~\ref{eq:cost}) as a dense tensor of $|\mathcal{A}|$ rows, each of length $\max_{r_i \in \mathcal{A}}(p_i + (b_i + 1) \times L)$.
To preserve positional semantics, we apply left-padding so that each prompt's tokens occupy contiguous positions ending immediately before its generation region.
We then compute explicit position IDs via cumulative indexing over the attention mask, ensuring that the first non-padding token always starts at logical position zero.
This preserves the logical relative distances required by rotary embeddings (RoPE)~\cite{su2024roformer} while keeping positional encoding, attention masking, and prefix-cache alignment consistent for each request within the batched execution.

\paragraph{Vectorized mixed-schedule execution}
While existing dLLM engines optimize single-request latency via dual cache and parallel decoding~\cite{wu2025fast}, applying these to mixed-state continuous batching is non-trivial because requests in a batch may be processing blocks at different sequence positions. BlockServe addresses this through gather-scatter indexing that selects each request's active block region within the shared batch tensor.

Standard dual cache mechanisms rely on static physical positions, which break under the dynamic realignments of continuous batching. We introduce a gather-scatter update mechanism that decouples logical token positions from physical memory addresses. By computing a sparse mask of active block regions, BlockServe performs in-place KV updates only at the physical offsets corresponding to the current execution block of each request, preserving the integrity of the frozen prefix cache across batch reconfigurations.
Standard parallel decoding requires all requests to be synchronized at the same block index. BlockServe removes this constraint by decoupling each request's generation trajectory. At the start of each block-cycle, the per-step transfer quota is computed by uniformly distributing the block's masked tokens across $S$ diffusion steps; since every block begins fully masked, this computation is block-agnostic. The same gather-based indexing enables a unified block-cycle to advance all active requests regardless of their individual block progress.

Gather-scatter indexing is the implementation primitive that makes mixed-state batching practical: gather extracts each request's active block region into a compact tensor for denoising, and scatter writes results back to the corresponding positions. This mechanism is not an independent algorithmic contribution but the execution substrate that enables block-grained scheduling over heterogeneous request states.

\section{Experiments}
\label{sec:experiments}

We evaluate BlockServe by answering the following research questions:
\begin{enumerate}
    \item \textbf{RQ1:} How does BlockServe compare to state-of-the-art baselines in terms of end-to-end throughput, total wall-clock time, and generation quality?
    \item \textbf{RQ2:} How does per-sample latency relate to output length under batching?
    \item \textbf{RQ3:} How does the system scale with increasing batch size?
    \item \textbf{RQ4:} What is the impact of token-budget admission control on batch capacity?
    \item \textbf{RQ5:} How do block length and prompt-length sorting affect performance?
\end{enumerate}

\subsection{Experimental setup}
We implement BlockServe and all baselines on a single NVIDIA H200 GPU using the LLaDA-8B-Instruct and Dream-v0-Instruct-7B models in bfloat16. The default block length is $L=32$ tokens with $S=32$ denoising steps per block.

We compare against two baselines: Vanilla, the default LLaDA and Dream configurations without serving optimizations, and Fast-dLLM~\cite{wu2025fast}, a dLLM inference accelerator with Dual Cache and Parallel Decoding enabled. For a fair comparison, BlockServe also incorporates these optimizations, isolating gains to our scheduling and memory management. We evaluate two BlockServe variants: BlockServe~(Fixed), which replaces the token-budget admission rule with a fixed batch-size cap, and BlockServe~(Budget), which uses the token budget policy to determine effective concurrency dynamically (Section~\ref{sec:method:admission}).

We report throughput (tokens/sec), total wall-clock time, and task-specific accuracy across five widely used standard benchmarks---GSM8K~\cite{cobbe2021training}, HumanEval~\cite{chen2021evaluating}, MBPP~\cite{austin2021program}, MATH~\cite{hendrycks2021measuring}, and TruthfulQA-Gen~\cite{lin2022truthfulqa}---covering mathematical reasoning, code generation, and factual question answering. Implementation details and benchmark statistics, including the gap between generation limits and actual output lengths, are provided in Section~\ref{app:setup}.

\subsubsection{Detailed experimental setup}
\label{app:setup}

\paragraph{Implementation details}
We implement BlockServe and all baselines on a single NVIDIA H200 GPU (141GB HBM3e) using the official model implementations of LLaDA-8B-Instruct and Dream-v0-Instruct-7B, both loaded in bfloat16 precision. The default block length is $L=32$ tokens with $S=32$ denoising steps per block. To ensure deterministic evaluation, all evaluations are conducted with \texttt{temperature=0.0} (greedy decoding) under fixed seeds. For BlockServe (Budget), the token budget is set based on offline profiling on the target H200 GPU. We use 5-shot prompting for GSM8K, 3-shot for MBPP, 4-shot for MATH, and 0-shot for HumanEval and TruthfulQA-Gen. All benchmarks use the test split.

\paragraph{Benchmarks}
We evaluate our system using five widely used standard benchmarks that together cover a diverse range of reasoning and generation tasks:
\begin{itemize}
    \item \textbf{GSM8K:} A dataset of grade-school math word problems~\cite{cobbe2021training}.
    \item \textbf{HumanEval:} A code-generation benchmark of 164 handwritten Python function-synthesis problems evaluated with unit tests~\cite{chen2021evaluating}.
    \item \textbf{MBPP:} A benchmark of programming tasks for synthesizing short Python programs from natural language descriptions~\cite{austin2021program}.
    \item \textbf{MATH:} A dataset of 12,500 challenging competition mathematics problems with step-by-step solutions~\cite{hendrycks2021measuring}.
    \item \textbf{TruthfulQA-Gen:} The generation setting of TruthfulQA, a benchmark to measure whether a language model is truthful in generating answers to questions~\cite{lin2022truthfulqa}. We adopt the reference-guided LLM-as-judge evaluation protocol of Badshah and Sajjad~\cite{badshah2025reference} for automated accuracy assessment.
\end{itemize}

\paragraph{Baselines}
We compare BlockServe against the following baselines:
\begin{itemize}
    \item \textbf{Vanilla:} The default LLaDA and Dream configurations, used as plain reference baselines in our evaluation without serving optimizations.
    \item \textbf{Fast-dLLM (Static)~\cite{wu2025fast}:} An inference acceleration method for dLLMs. Fast-dLLM does not include a batching mechanism; in our evaluation, requests are processed in fixed groups where all must finish before the next batch begins. To ensure a strong baseline, we enable Dual Cache and Parallel Decoding.
    \item \textbf{BlockServe (Ours):} Our proposed system equipped with block-grained mixed-state scheduling. We evaluate two admission policies: BlockServe (Fixed) with a fixed batch size limit, and BlockServe (Budget) with the token budget policy. Both variants employ the length-aware sorting strategy (Section~\ref{sec:method:admission}) to minimize padding. For fair comparison, BlockServe also incorporates Dual Cache and Parallel Decoding, ensuring that performance gains are attributed to our scheduling and memory management rather than kernel-level optimizations.
\end{itemize}

\paragraph{Metrics}
We evaluate performance using the following metrics:
\begin{itemize}
    \item \textbf{Accuracy:} Task-specific accuracy to assess whether generation quality remains comparable across BlockServe and the baselines.
    \item \textbf{Throughput:} Defined as the total number of valid generated tokens (excluding padding and masked tokens) divided by the total wall-clock time (in seconds). This measures the effective computation rate of the serving system.
    \item \textbf{Total Time:} The wall-clock time required to complete the entire benchmark dataset.
\end{itemize}

\paragraph{Benchmark statistics}
\label{app:benchmark_stats}
We provide a comprehensive breakdown of the benchmark datasets used in our evaluation in Table~\ref{tab:benchmark_stats}. The substantial variance between the predefined generation limit and the actual generated sequence length---most notably observed in the TruthfulQA benchmark---underscores the inefficiency of static batching mechanisms. Such systems are compelled to reserve memory resources based on the worst-case generation limit, leading to significant underutilization when actual generations are short.

\begin{table*}[t]
\centering
\small
\caption{Evaluation benchmark statistics (LLaDA tokenizer).}
\label{tab:benchmark_stats}
\begin{tabular}{lcccc}
\toprule
\textbf{Dataset} & \textbf{Shots} & \textbf{Prompt Len (Avg/Max)} & \textbf{Gen Limit} & \textbf{Actual Gen Len (Avg/Max)} \\
\midrule
\multirow{2}{*}{\textbf{GSM8K}} & \multirow{2}{*}{5-shot} & \multirow{2}{*}{1020 / 1632} & 256 & 231 / 260 \\
 & & & 512 & 265 / 511 \\
\midrule
\multirow{2}{*}{\textbf{HumanEval}} & \multirow{2}{*}{0-shot} & \multirow{2}{*}{157 / 423} & 256 & 231 / 255 \\
 & & & 512 & 466 / 512 \\
\midrule
\multirow{2}{*}{\textbf{MBPP}} & \multirow{2}{*}{3-shot} & \multirow{2}{*}{750 / 4332} & 256 & 148 / 255 \\
 & & & 512 & 284 / 512 \\
\midrule
\multirow{2}{*}{\textbf{MATH}} & \multirow{2}{*}{4-shot} & \multirow{2}{*}{698 / 2029} & 512 & 428 / 514 \\
 & & & 1024 & 600 / 1027 \\
\midrule
\multirow{3}{*}{\textbf{TruthfulQA}} & \multirow{3}{*}{0-shot} & \multirow{3}{*}{198 / 245} & 256 & 33 / 254 \\
 & & & 512 & 33 / 510 \\
 & & & 1024 & 36 / 1022 \\
\bottomrule
\end{tabular}
\end{table*}

\subsection{End-to-end performance (RQ1)}
We evaluate the end-to-end performance in terms of tokens per second (TPS), accuracy, and total wall-clock time. Table~\ref{tab:main_results_3_row} summarizes the results for LLaDA; Dream results, which exhibit consistent trends, are provided in Section~\ref{app:dream_results}.

\paragraph{Throughput improvement}
BlockServe consistently achieves higher throughput than the Fast-dLLM baseline across all benchmarks. For example, on HumanEval with LLaDA (Table~\ref{tab:main_results_3_row}, Gen Len 256), BlockServe increases throughput from 231.5 TPS (Fast-dLLM, best-performing batch size) to 552.6 TPS (BlockServe, Fixed, batch size (BS)=16), and further reaches 597.8 TPS with token-budget admission control (Budget, reported as the peak throughput across batch sizes), corresponding to a 2.6$\times$ speedup over the best-performing Fast-dLLM configuration we evaluated. On LLaDA, speedups range from 2.6$\times$ to 7.7$\times$ across benchmarks (Table~\ref{tab:main_results_3_row}); on Dream, the range extends to 1.9--10.6$\times$ (Section~\ref{app:dream_results}).

\paragraph{Accuracy maintenance}
Although all evaluated methods exhibit minor accuracy variations under different batch size settings, BlockServe achieves accuracy comparable to the baselines across the benchmarks considered. As shown in Table~\ref{tab:main_results_3_row}, BlockServe maintains generation quality broadly comparable to the baselines. For instance, on the MATH benchmark, BlockServe matches the baseline accuracy of $\sim$36\%, suggesting that our throughput gains are achieved while maintaining comparable generation accuracy.
\begin{table*}[t]
\centering
\normalsize
\caption{Performance comparison on LLaDA (batch sizes 4, 8, 16). TPS: Tokens/s. Acc: Accuracy (\%). Time: Wall-clock time. Dash: exceeded 24h or OOM. Best TPS in bold.}
\label{tab:main_results_3_row}
\resizebox{\linewidth}{!}{
\begin{tabular}{lcccccccccccccc}
\toprule
\multirow{2}{*}{\textbf{Dataset}} & \textbf{Gen} & \textbf{Batch} & \multicolumn{3}{c}{\textbf{LLaDA (Default)}} & \multicolumn{3}{c}{\textbf{Fast-dLLM}} & \multicolumn{3}{c}{\textbf{BlockServe (Fixed)}} & \multicolumn{3}{c}{\textbf{BlockServe (Budget)}} \\
\cmidrule(lr){4-6} \cmidrule(lr){7-9} \cmidrule(lr){10-12} \cmidrule(lr){13-15}
 & \textbf{Len} & \textbf{Size} & \textbf{TPS} & \textbf{Acc} & \textbf{Time} & \textbf{TPS} & \textbf{Acc} & \textbf{Time} & \textbf{TPS} & \textbf{Acc} & \textbf{Time} & \textbf{TPS} & \textbf{Acc} & \textbf{Time} \\
\midrule
\multirow{6}{*}{\textbf{GSM8K}} & \multirow{3}{*}{256} & 4 & 17.5 & 77.71 & 17547.4 s & 145.6 & 78.39 & 2098.4 s & 248.8 & 77.56 & 1231.6 s & \multirow{3}{*}{\textbf{392.5}} & \multirow{3}{*}{77.86} & \multirow{3}{*}{778.2 s} \\
 &  & 8 & 17.0 & 77.41 & 18094.9 s & 149.2 & 78.09 & 2054.3 s & 330.4 & 77.48 & 925.6 s &  &  &  \\
 &  & 16 & 16.4 & 76.95 & 18842.5 s & 142.3 & 77.94 & 2158.3 s & 378.9 & 78.85 & 807.0 s &  &  &  \\
\cmidrule{2-15}
 & \multirow{3}{*}{512} & 4 & 8.5 & 76.72 & 41469.7 s & 102.4 & 74.68 & 3458.8 s & 229.0 & 75.06 & 1535.7 s & \multirow{3}{*}{\textbf{334.7}} & \multirow{3}{*}{76.12} & \multirow{3}{*}{1038.8 s} \\
 &  & 8 & 8.3 & 77.41 & 42630.7 s & 95.0 & 75.06 & 3715.8 s & 292.4 & 75.51 & 1208.4 s &  &  &  \\
 &  & 16 & 8.0 & 76.72 & 44295.1 s & 83.9 & 74.91 & 4205.4 s & 325.6 & 76.65 & 1072.0 s &  &  &  \\
\midrule
\multirow{6}{*}{\textbf{HumanEval}} & \multirow{3}{*}{256} & 4 & 53.5 & 41.46 & 736.8 s & 193.1 & 32.32 & 197.8 s & 284.9 & 38.41 & 135.8 s & \multirow{3}{*}{\textbf{597.8}} & \multirow{3}{*}{38.41} & \multirow{3}{*}{63.4 s} \\
 &  & 8 & 51.8 & 39.63 & 761.0 s & 226.2 & 31.10 & 169.8 s & 409.8 & 38.41 & 93.4 s &  &  &  \\
 &  & 16 & 48.9 & 39.63 & 798.8 s & 231.5 & 34.15 & 165.9 s & 552.6 & 36.59 & 67.8 s &  &  &  \\
\cmidrule{2-15}
 & \multirow{3}{*}{512} & 4 & 34.2 & 45.73 & 2226.1 s & 184.4 & 45.12 & 410.9 s & 276.0 & 43.29 & 279.1 s & \multirow{3}{*}{\textbf{538.1}} & \multirow{3}{*}{43.29} & \multirow{3}{*}{140.1 s} \\
 &  & 8 & 33.9 & 48.78 & 2253.4 s & 195.1 & 44.51 & 389.0 s & 401.7 & 42.68 & 190.6 s &  &  &  \\
 &  & 16 & 32.7 & 48.78 & 2339.6 s & 197.1 & 43.90 & 385.7 s & 509.8 & 45.12 & 150.0 s &  &  &  \\
\midrule
\multirow{6}{*}{\textbf{MBPP}} & \multirow{3}{*}{256} & 4 & 15.9 & 32.00 & 5064.1 s & 105.1 & 29.00 & 662.4 s & 219.3 & 25.80 & 345.0 s & \multirow{3}{*}{\textbf{363.1}} & \multirow{3}{*}{27.00} & \multirow{3}{*}{190.5 s} \\
 &  & 8 & 15.5 & 31.60 & 5298.2 s & 99.8 & 29.00 & 705.4 s & 302.2 & 26.20 & 250.4 s &  &  &  \\
 &  & 16 & - & - & - & 92.5 & 29.00 & 758.1 s & 344.5 & 27.00 & 210.8 s &  &  &  \\
\cmidrule{2-15}
 & \multirow{3}{*}{512} & 4 & 10.4 & 20.60 & 12505.3 s & 93.2 & 17.60 & 1388.2 s & 220.9 & 14.40 & 655.9 s & \multirow{3}{*}{\textbf{334.5}} & \multirow{3}{*}{15.60} & \multirow{3}{*}{413.5 s} \\
 &  & 8 & 9.7 & 22.40 & 12965.8 s & 83.5 & 19.40 & 1486.0 s & 290.0 & 14.40 & 483.7 s &  &  &  \\
 &  & 16 & - & - & - & 77.5 & 19.60 & 1591.8 s & 321.5 & 14.20 & 423.6 s &  &  &  \\
\midrule
\multirow{6}{*}{\textbf{MATH}} & \multirow{3}{*}{512} & 4 & - & - & \textgreater 24 h & 138.7 & 35.94 & 15380.2 s & 252.3 & 36.28 & 8499.6 s & \multirow{3}{*}{\textbf{401.0}} & \multirow{3}{*}{36.36} & \multirow{3}{*}{5358.6 s} \\
 &  & 8 & - & - & \textgreater 24 h & 141.4 & 35.46 & 15081.6 s & 329.7 & 35.80 & 6510.7 s &  &  &  \\
 &  & 16 & - & - & \textgreater 24 h & 135.2 & 35.20 & 15779.3 s & 386.8 & 36.24 & 5547.5 s &  &  &  \\
\cmidrule{2-15}
 & \multirow{3}{*}{1024} & 4 & - & - & \textgreater 24 h & 103.2 & 35.02 & 28866.5 s & 221.5 & 35.76 & 13649.9 s & \multirow{3}{*}{\textbf{314.3}} & \multirow{3}{*}{35.30} & \multirow{3}{*}{9563.6 s} \\
 &  & 8 & - & - & \textgreater 24 h & 96.6 & 34.68 & 30666.6 s & 278.4 & 35.14 & 10823.8 s &  &  &  \\
 &  & 16 & - & - & \textgreater 24 h & 86.8 & 34.66 & 34144.3 s & 309.8 & 35.04 & 9713.5 s &  &  &  \\
\midrule
\multirow{9}{*}{\textbf{TruthfulQA}} & \multirow{3}{*}{256} & 4 & 7.4 & 40.88 & 3470.2 s & 35.6 & 39.78 & 766.4 s & 78.6 & 38.56 & 352.6 s & \multirow{3}{*}{\textbf{217.6}} & \multirow{3}{*}{37.94} & \multirow{3}{*}{127.4 s} \\
 &  & 8 & 7.8 & 40.51 & 3370.9 s & 38.2 & 37.82 & 700.0 s & 132.1 & 39.41 & 210.8 s &  &  &  \\
 &  & 16 & 7.8 & 40.88 & 3312.9 s & 37.0 & 37.33 & 726.8 s & 187.3 & 38.19 & 148.3 s &  &  &  \\
\cmidrule{2-15}
 & \multirow{3}{*}{512} & 4 & 2.5 & 39.41 & 10663.9 s & 22.0 & 37.21 & 1227.6 s & 55.5 & 36.84 & 500.4 s & \multirow{3}{*}{\textbf{140.9}} & \multirow{3}{*}{37.82} & \multirow{3}{*}{200.2 s} \\
 &  & 8 & 2.6 & 38.56 & 10442.8 s & 20.4 & 36.60 & 1337.1 s & 89.2 & 36.96 & 308.6 s &  &  &  \\
 &  & 16 & 2.6 & 39.41 & 10249.5 s & 18.8 & 37.09 & 1455.8 s & 126.3 & 37.09 & 222.4 s &  &  &  \\
\cmidrule{2-15}
 & \multirow{3}{*}{1024} & 4 & 0.8 & 39.53 & 36442.5 s & 14.6 & 36.60 & 2033.0 s & 54.0 & 36.60 & 552.3 s & \multirow{3}{*}{\textbf{112.8}} & \multirow{3}{*}{35.25} & \multirow{3}{*}{264.4 s} \\
 &  & 8 & 0.8 & 38.92 & 35788.0 s & 13.3 & 35.74 & 2118.6 s & 81.7 & 36.60 & 376.2 s &  &  &  \\
 &  & 16 & 0.8 & 39.29 & 35765.5 s & 13.2 & 35.86 & 2287.3 s & 101.7 & 35.37 & 313.2 s &  &  &  \\
\bottomrule
\end{tabular}
}
\end{table*}

\subsubsection{Dream model results (RQ1)}
\label{app:dream_results}
Table~\ref{tab:dream_results_3_row} reports the end-to-end results for the Dream model across all five benchmarks. The trends mirror those observed for LLaDA in Section~\ref{sec:experiments}: BlockServe achieves 1.9--10.6$\times$ throughput improvement over Fast-dLLM across benchmarks while maintaining comparable generation accuracy. The largest speedup (10.6$\times$) occurs on TruthfulQA at generation length 1024, where short output lengths amplify the straggler effect in static batching.

\begin{table*}[t]
\centering
\normalsize
\caption{Performance comparison on Dream (batch sizes 4, 8, 16). TPS: Tokens/s. Acc: Accuracy (\%). Time: Wall-clock time. Dash: exceeded 24h or OOM. Best TPS in bold.}
\label{tab:dream_results_3_row}
\resizebox{\linewidth}{!}{
\begin{tabular}{lcccccccccccccc}
\toprule
\multirow{2}{*}{\textbf{Dataset}} & \textbf{Gen} & \textbf{Batch} & \multicolumn{3}{c}{\textbf{Dream (Default)}} & \multicolumn{3}{c}{\textbf{Fast-dLLM}} & \multicolumn{3}{c}{\textbf{BlockServe (Fixed)}} & \multicolumn{3}{c}{\textbf{BlockServe (Budget)}} \\
\cmidrule(lr){4-6} \cmidrule(lr){7-9} \cmidrule(lr){10-12} \cmidrule(lr){13-15}
 & \textbf{Len} & \textbf{Size} & \textbf{TPS} & \textbf{Acc} & \textbf{Time} & \textbf{TPS} & \textbf{Acc} & \textbf{Time} & \textbf{TPS} & \textbf{Acc} & \textbf{Time} & \textbf{TPS} & \textbf{Acc} & \textbf{Time} \\
\midrule
\multirow{6}{*}{\textbf{GSM8K}} & \multirow{3}{*}{256} & 4 & 5.1 & 45.19 & 16112.1 s & 124.2 & 77.18 & 1405.1 s & 222.9 & 76.50 & 782.0 s & \multirow{3}{*}{\textbf{280.3}} & \multirow{3}{*}{76.35} & \multirow{3}{*}{620.9 s} \\
 &  & 8 & 5.0 & 44.88 & 16646.1 s & 142.1 & 77.10 & 1229.6 s & 263.3 & 74.68 & 661.7 s &  &  &  \\
 &  & 16 & 4.8 & 45.26 & 17183.3 s & 147.2 & 76.42 & 1183.8 s & 279.0 & 76.19 & 626.2 s &  &  &  \\
\cmidrule{2-15}
 & \multirow{3}{*}{512} & 4 & 2.5 & 47.01 & 38642.9 s & 77.7 & 75.36 & 2091.4 s & 199.6 & 74.53 & 810.4 s & \multirow{3}{*}{\textbf{238.5}} & \multirow{3}{*}{76.42} & \multirow{3}{*}{690.2 s} \\
 &  & 8 & 2.5 & 47.31 & 39665.7 s & 84.5 & 76.50 & 1944.1 s & 228.6 & 74.83 & 714.1 s &  &  &  \\
 &  & 16 & 2.4 & 47.92 & 40968.7 s & 83.9 & 75.66 & 1953.6 s & 238.1 & 75.97 & 693.0 s &  &  &  \\
\midrule
\multirow{6}{*}{\textbf{HumanEval}} & \multirow{3}{*}{256} & 4 & 5.4 & 20.12 & 633.6 s & 130.9 & 57.93 & 157.1 s & 293.9 & 57.32 & 71.3 s & \multirow{3}{*}{\textbf{477.9}} & \multirow{3}{*}{57.32} & \multirow{3}{*}{44.6 s} \\
 &  & 8 & 5.4 & 20.12 & 646.6 s & 183.8 & 59.15 & 111.8 s & 407.3 & 58.54 & 53.1 s &  &  &  \\
 &  & 16 & 5.0 & 19.51 & 673.2 s & 205.1 & 58.54 & 98.7 s & 462.9 & 56.71 & 45.2 s &  &  &  \\
\cmidrule{2-15}
 & \multirow{3}{*}{512} & 4 & 1.7 & 18.29 & 1979.9 s & 108.9 & 57.32 & 227.3 s & 252.8 & 56.10 & 103.8 s & \multirow{3}{*}{\textbf{374.8}} & \multirow{3}{*}{60.98} & \multirow{3}{*}{66.2 s} \\
 &  & 8 & 1.7 & 18.29 & 1999.4 s & 130.2 & 57.93 & 186.6 s & 328.8 & 57.32 & 78.5 s &  &  &  \\
 &  & 16 & 1.7 & 19.51 & 2054.1 s & 142.2 & 56.71 & 176.4 s & 369.4 & 59.15 & 68.0 s &  &  &  \\
\midrule
\multirow{6}{*}{\textbf{MBPP}} & \multirow{3}{*}{256} & 4 & 2.1 & 26.20 & 4420.4 s & 83.5 & 53.00 & 292.9 s & 186.0 & 52.60 & 126.8 s & \multirow{3}{*}{\textbf{220.8}} & \multirow{3}{*}{50.80} & \multirow{3}{*}{105.9 s} \\
 &  & 8 & 2.0 & 26.60 & 4646.6 s & 91.6 & 52.00 & 265.1 s & 209.9 & 51.40 & 112.3 s &  &  &  \\
 &  & 16 & 1.8 & 26.60 & 5057.8 s & 89.5 & 52.20 & 274.9 s & 195.3 & 51.00 & 119.7 s &  &  &  \\
\cmidrule{2-15}
 & \multirow{3}{*}{512} & 4 & 0.9 & 29.80 & 11155.8 s & 46.6 & 51.60 & 532.4 s & 162.6 & 52.00 & 144.1 s & \multirow{3}{*}{\textbf{201.0}} & \multirow{3}{*}{51.60} & \multirow{3}{*}{118.7 s} \\
 &  & 8 & 0.9 & 30.20 & 11601.9 s & 49.3 & 52.00 & 505.6 s & 191.2 & 51.80 & 123.8 s &  &  &  \\
 &  & 16 & 0.8 & 28.80 & 12348.7 s & 48.9 & 51.60 & 518.2 s & 182.9 & 51.60 & 128.2 s &  &  &  \\
\midrule
\multirow{6}{*}{\textbf{MATH}} & \multirow{3}{*}{512} & 4 & - & - & \textgreater 24 h & 122.8 & 37.76 & 9337.5 s & 251.8 & 37.18 & 4545.2 s & \multirow{3}{*}{\textbf{305.1}} & \multirow{3}{*}{37.28} & \multirow{3}{*}{3733.6 s} \\
 &  & 8 & - & - & \textgreater 24 h & 142.5 & 37.38 & 8063.4 s & 292.5 & 36.86 & 3933.7 s &  &  &  \\
 &  & 16 & - & - & \textgreater 24 h & 146.9 & 37.58 & 7766.0 s & 305.4 & 37.26 & 3725.4 s &  &  &  \\
\cmidrule{2-15}
 & \multirow{3}{*}{1024} & 4 & - & - & \textgreater 24 h & 83.8 & 37.18 & 17060.0 s & 220.4 & 37.26 & 6533.7 s & \multirow{3}{*}{\textbf{234.0}} & \multirow{3}{*}{37.70} & \multirow{3}{*}{6097.4 s} \\
 &  & 8 & - & - & \textgreater 24 h & 89.5 & 37.30 & 15827.6 s & 235.8 & 36.94 & 6061.6 s &  &  &  \\
 &  & 16 & - & - & \textgreater 24 h & 88.6 & 37.30 & 16139.6 s & 235.6 & 37.52 & 6089.5 s &  &  &  \\
\midrule
\multirow{9}{*}{\textbf{TruthfulQA}} & \multirow{3}{*}{256} & 4 & 1.3 & 42.96 & 2961.5 s & 35.1 & 51.90 & 262.6 s & 96.3 & 49.08 & 86.7 s & \multirow{3}{*}{\textbf{174.9}} & \multirow{3}{*}{48.84} & \multirow{3}{*}{48.2 s} \\
 &  & 8 & 1.3 & 42.59 & 2916.7 s & 46.2 & 50.55 & 203.7 s & 133.6 & 48.96 & 62.7 s &  &  &  \\
 &  & 16 & 1.3 & 42.96 & 2865.0 s & 51.5 & 49.94 & 180.9 s & 160.8 & 48.35 & 52.7 s &  &  &  \\
\cmidrule{2-15}
 & \multirow{3}{*}{512} & 4 & 0.4 & 43.45 & 9447.7 s & 20.4 & 52.63 & 510.0 s & 89.3 & 52.14 & 100.3 s & \multirow{3}{*}{\textbf{146.9}} & \multirow{3}{*}{51.77} & \multirow{3}{*}{61.1 s} \\
 &  & 8 & 0.4 & 43.57 & 9255.6 s & 24.1 & 52.14 & 424.2 s & 116.6 & 52.51 & 75.6 s &  &  &  \\
 &  & 16 & 0.4 & 44.06 & 9202.8 s & 25.4 & 52.39 & 396.2 s & 135.2 & 51.77 & 65.9 s &  &  &  \\
\cmidrule{2-15}
 & \multirow{3}{*}{1024} & 4 & 0.1 & 42.35 & 33687.9 s & 8.2 & 54.83 & 1317.7 s & 70.2 & 52.88 & 127.7 s & \multirow{3}{*}{\textbf{99.0}} & \multirow{3}{*}{52.39} & \multirow{3}{*}{91.3 s} \\
 &  & 8 & 0.1 & 41.98 & 33138.4 s & 9.0 & 54.71 & 1206.6 s & 85.9 & 53.86 & 104.0 s &  &  &  \\
 &  & 16 & 0.1 & 42.59 & 32371.4 s & 9.3 & 54.59 & 1166.9 s & 94.3 & 52.02 & 95.4 s &  &  &  \\
\bottomrule
\end{tabular}
}
\end{table*}

\subsection{Latency--length relationship (RQ2)}
\begin{figure*}[t]
    \centering
    \subfloat[BS=4\label{fig:dream_violin_bs4}]{%
        \includegraphics[width=0.47\linewidth]{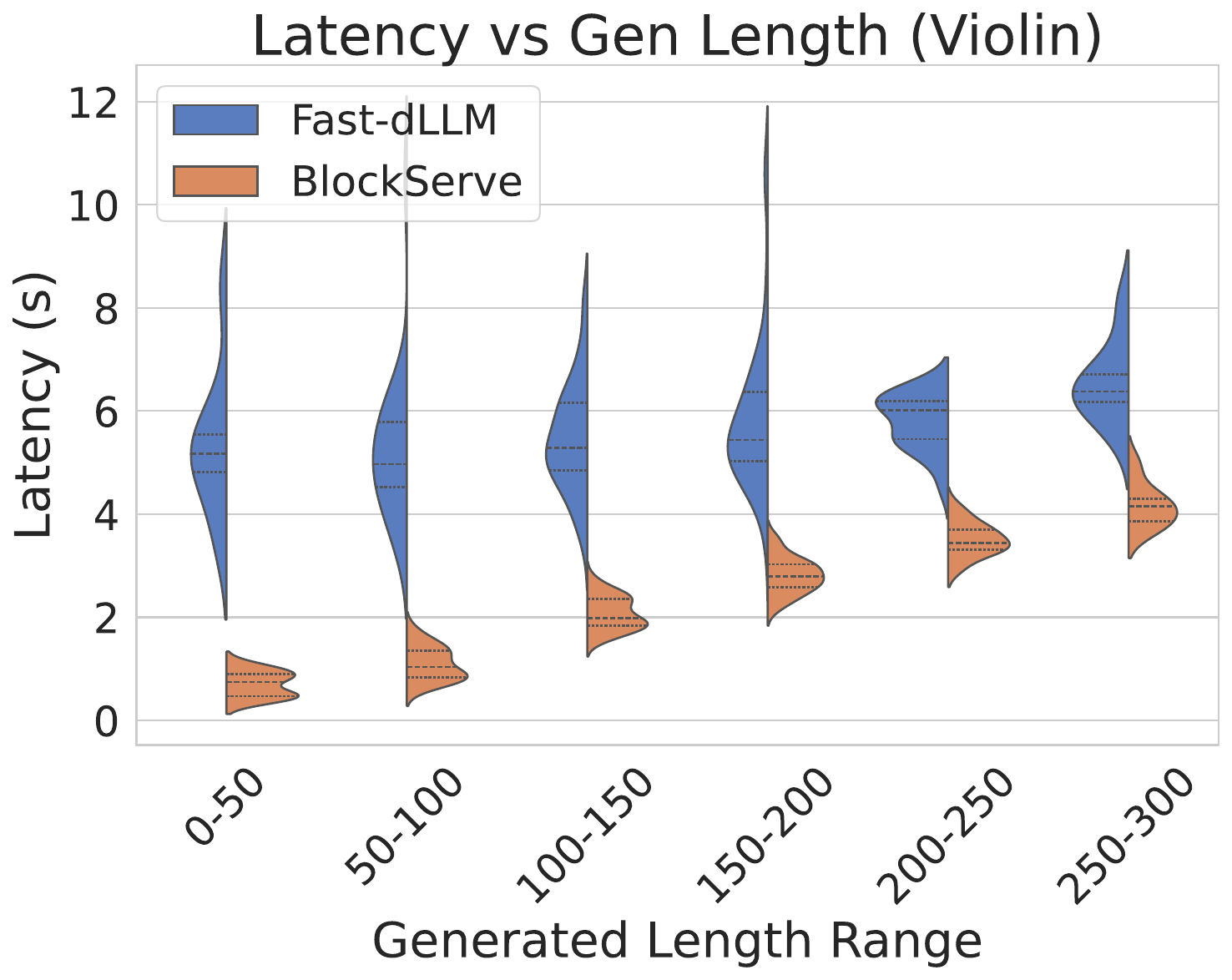}}
    \hfill
    \subfloat[BS=16\label{fig:dream_violin_bs16}]{%
        \includegraphics[width=0.47\linewidth]{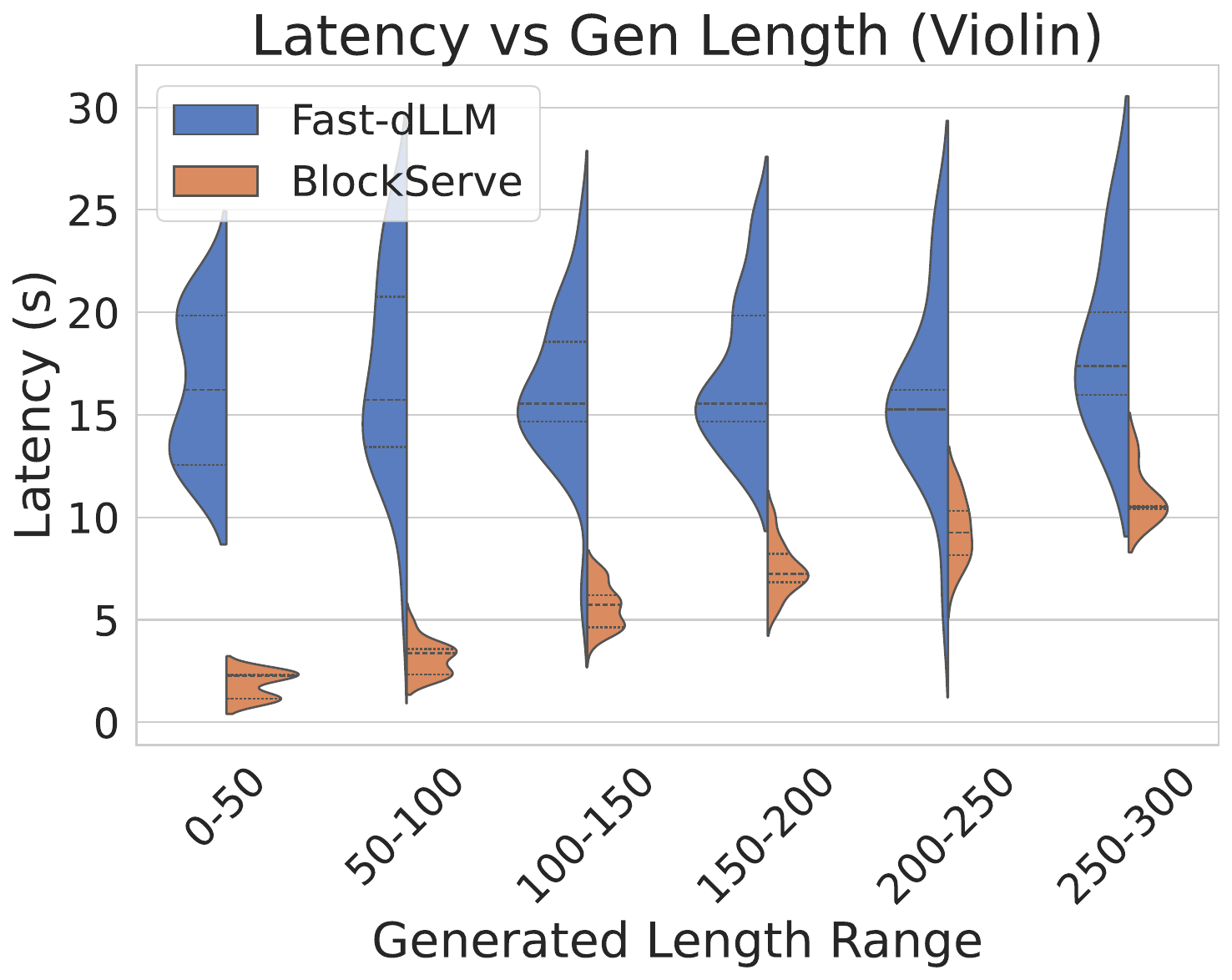}}
    \caption{Dream latency distribution (HumanEval, gen len 512). Violin plots show per-sample latency density at each output-length bucket.}
    \label{fig:dream_latency_dist}
\end{figure*}

To understand the source of our speedup, we analyze per-sample service-time latency---the wall-clock time from scheduling to completion---as a function of output length. Figure~\ref{fig:dream_latency_dist} presents the distribution for Dream on HumanEval at batch sizes 4 and 16.

\paragraph{Latency analysis}
Figure~\ref{fig:dream_latency_dist} highlights the distributional advantage of BlockServe. BlockServe maintains a consistent, low-variance latency across output-length buckets. In contrast, Fast-dLLM exhibits pronounced ``straggler effects'' characterized by a heavy tail and multiple modes. Even for short generations, the baseline incurs high latency because the entire batch remains blocked by the longest-running request.

\begin{figure*}[t]
    \centering
    \subfloat[TruthfulQA (Gen Len 512).\label{fig:tqa_bubbles}]{%
        \includegraphics[width=0.47\linewidth]{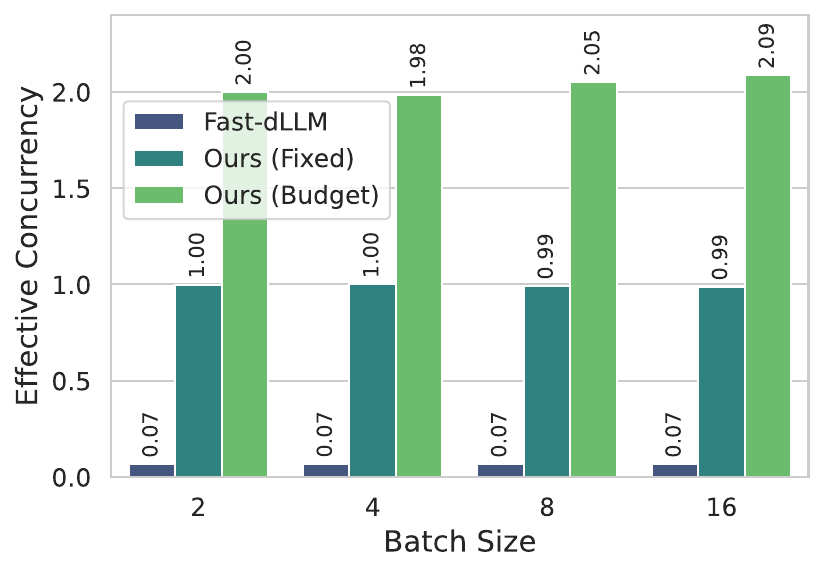}}
    \hfill
    \subfloat[HumanEval (Gen Len 512).\label{fig:he_bubbles}]{%
        \includegraphics[width=0.47\linewidth]{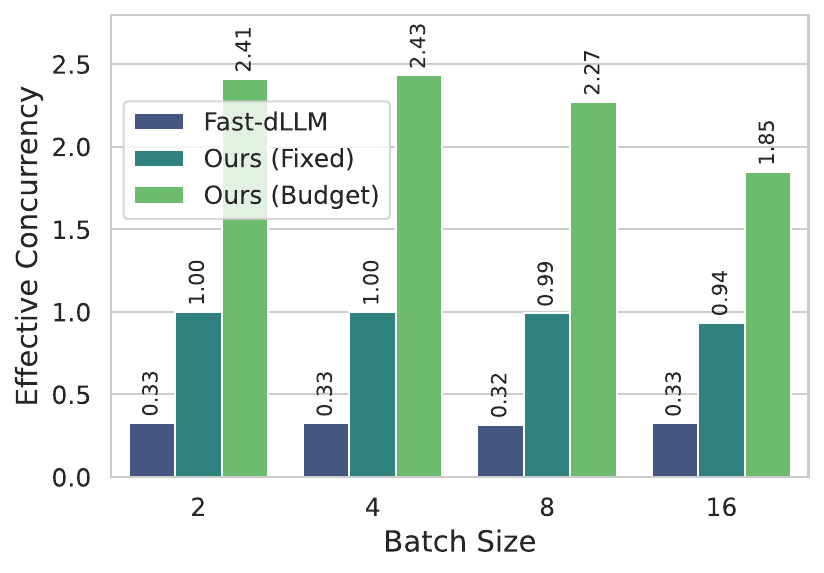}}
    \caption{Effective concurrency analysis via compute ratio (Eq.~\ref{eq:effective_compute_ratio}). Results shown for Dream.}
    \label{fig:utilization_analysis}
\end{figure*}

\paragraph{Effective concurrency analysis}
Figure~\ref{fig:utilization_analysis} quantifies the effective concurrency gap between methods.
We report the Effective Compute Ratio \(R\) (Eq.~\ref{eq:effective_compute_ratio}), measuring the average number of active requests per scheduling step, normalized by the baseline batch size:
\begin{equation}
\label{eq:effective_compute_ratio}
R = \frac{1}{TB}\sum_{t=1}^{T} a_t,
\end{equation}
where \(a_t\) is the number of active requests at block-cycle \(t\), \(B\) is the baseline batch size, and \(T\) is the total number of block-cycles.
For Fast-dLLM, the effective ratio drops sharply because finished requests occupy slots while the batch completes. On TruthfulQA (Figure~\ref{fig:tqa_bubbles}), which has short generations, the ratio at BS=16 drops to 0.07. In contrast, BlockServe (Fixed) maintains a ratio near 1.0, indicating that block-grained preemption effectively reduces compute bubbles. BlockServe (Budget) can exceed 2.0 (e.g., $\sim$2.4 at BS=4 on HumanEval, Figure~\ref{fig:he_bubbles}), reflecting efficient oversubscription under token-budget control.

\subsection{Batch-size scaling of straggler effects (RQ3)}

\begin{figure*}[t]
    \centering
    \subfloat[HumanEval (Gen Len 512)\label{fig:llada_scaling_he}]{%
        \includegraphics[width=0.47\linewidth]{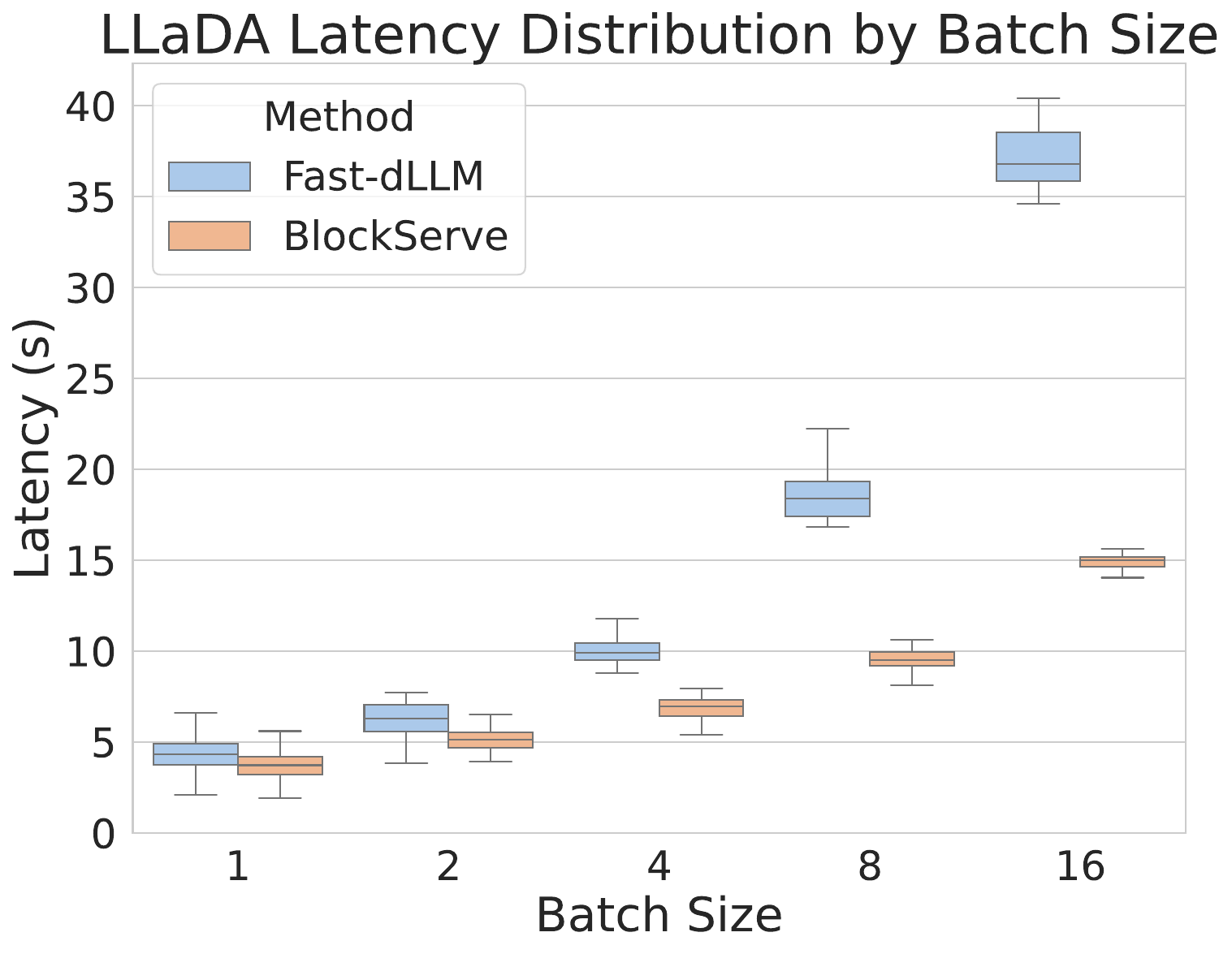}}
    \hfill
    \subfloat[TruthfulQA (Gen Len 512)\label{fig:llada_scaling_tqa}]{%
        \includegraphics[width=0.47\linewidth]{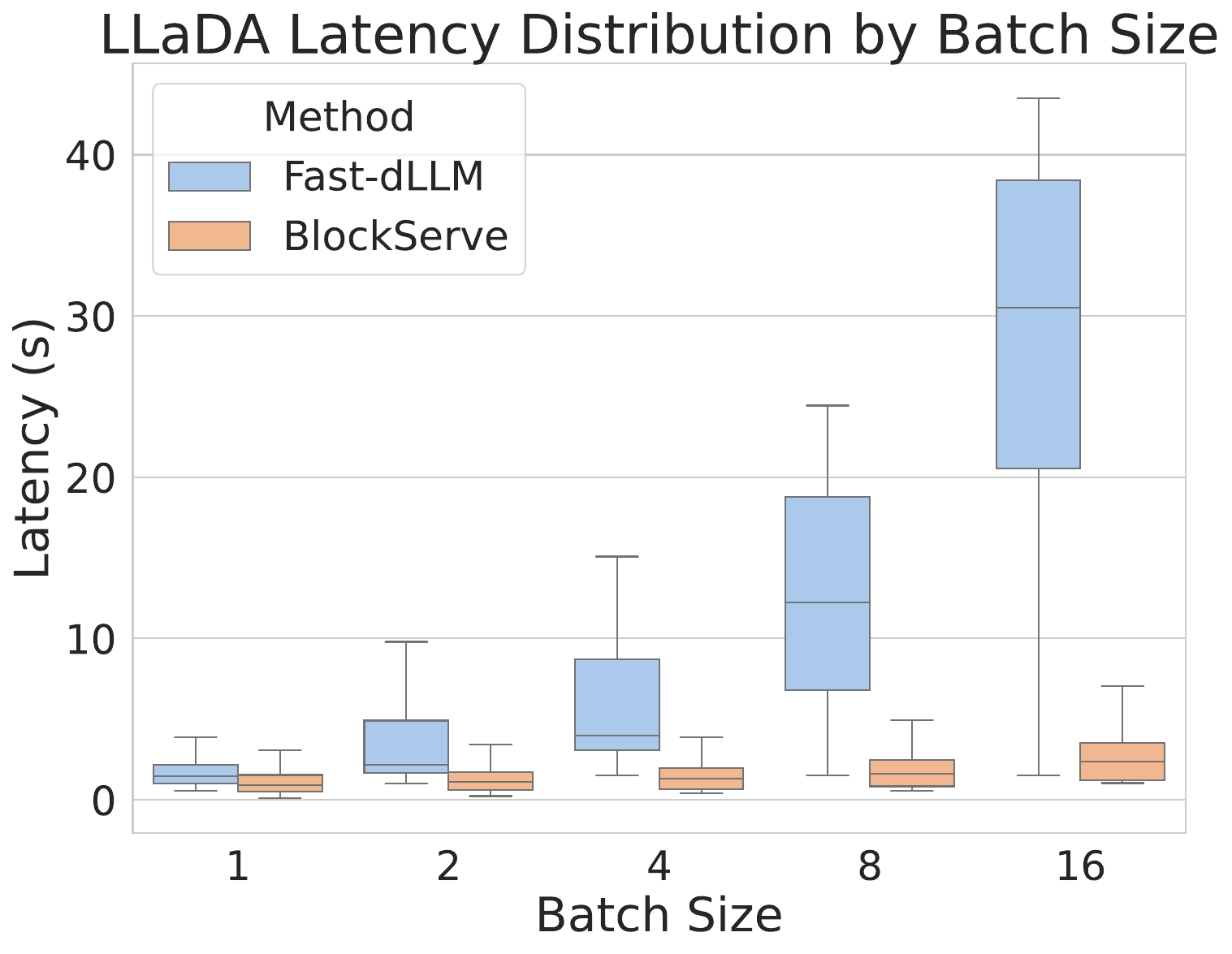}}
    \caption{Batch size scaling (LLaDA). Boxplots show per-sample latency for batch sizes 1--16.}
    \label{fig:llada_scaling}
\end{figure*}

Figure~\ref{fig:llada_scaling} demonstrates the critical scalability advantage of BlockServe. At small batch sizes (BS=1, 2), the latency gap between Fast-dLLM (Blue) and BlockServe (Orange) is minimal. However, as batch size increases to 16, the baseline latency grows disproportionately because the probability of encountering a long straggler request within the batch increases, leading to extreme tail latency, particularly on HumanEval where generation lengths vary significantly. BlockServe effectively decouples request latency from batch composition: for compute-intensive tasks like HumanEval, latency grows steadily with batch size; for shorter tasks like TruthfulQA, latency remains nearly constant as batch size increases. BlockServe's latency distribution remains compact even at BS=16, confirming that block-grained scheduling ensures predictable latency even under heavy contention.

\subsection{Admission control and batch capacity (RQ4)}
\label{app:rq4}
We compare the maximum safe batch size, defined as the largest number of concurrent requests that completes without triggering OOM under a fixed generation-length limit. Table~\ref{tab:max_batch_size} reports the per-benchmark results.
Under static batching, Fast-dLLM must reserve all memory slots for the entire batch lifetime, enforcing a conservative batch size.
BlockServe reclaims slots at block granularity and admits new requests under the token budget, progressively tightening the memory bounding box as high-progress requests complete (Section~\ref{sec:method:admission}). For example, on LLaDA MBPP, the maximum batch size increases from 16 (Fast-dLLM) to 64 (BlockServe), a $4.0\times$ improvement. Across benchmarks, this yields a $2.0\times$--$4.0\times$ increase in effective batch capacity under the same hardware constraints.

Table~\ref{tab:max_batch_size} compares the maximum safe batch size---the largest number of concurrent requests that completes without OOM under a fixed generation length of 1024 tokens. BlockServe's token-budget admission control achieves $2.0\times$--$4.0\times$ larger batch capacity than Fast-dLLM's fixed limit across both models and the evaluated benchmarks, confirming that block-grained slot reclamation effectively relaxes the conservative memory constraints of static batching.

\begin{table}[t]
\centering
\small
\caption{Maximum safe batch size comparison (generation length fixed at 1024 tokens). Fast-dLLM: fixed batch limit. BlockServe: token-budget peak capacity.}
\label{tab:max_batch_size}

\resizebox{\columnwidth}{!}{%
\begin{tabular}{llccc}
\toprule
\textbf{Model} & \textbf{Dataset} & \textbf{Fast-dLLM} & \textbf{BlockServe} & \textbf{Improvement} \\
\midrule
\multirow{4}{*}{\textbf{Dream}} & GSM8K & 32 & 64 & 2.0$\times$ \\
 & HumanEval & 64 & 128 & 2.0$\times$ \\
 & MBPP & 32 & 71 & 2.2$\times$ \\
 & TruthfulQA & 64 & 128 & 2.0$\times$ \\
\midrule
\multirow{4}{*}{\textbf{LLaDA}} & GSM8K & 16 & 37 & 2.3$\times$ \\
 & HumanEval & 32 & 64 & 2.0$\times$ \\
 & MBPP & 16 & 64 & 4.0$\times$ \\
 & TruthfulQA & 16 & 64 & 4.0$\times$ \\
\bottomrule
\end{tabular}
}
\end{table}

\subsection{Ablation studies (RQ5)}
\label{sec:ablation}

We ablate two design choices affecting throughput--quality trade-offs: block length $L$, which controls preemption granularity, and prompt-length sorting, which reduces padding waste.

\paragraph{Impact of block generation length}
We evaluate block length $L \in \{16, 32, 64\}$ on GSM8K and HumanEval. The results are presented in Table~\ref{tab:ablation_block_size_gsm8k} and Table~\ref{tab:ablation_block_size_humaneval}.

\begin{table*}[t]
    \centering
    \begin{minipage}[t]{0.48\linewidth}
        \centering
        \caption{Block length ($L$) ablation on GSM8K (Gen Len 256, Batch Size 4/8).}
        \label{tab:ablation_block_size_gsm8k}
        \resizebox{\linewidth}{!}{
        \begin{tabular}{lccccc}
        \toprule
        \multirow{2}{*}{\textbf{Model}} & \multirow{2}{*}{$L$} & \multicolumn{2}{c}{\textbf{BS=4}} & \multicolumn{2}{c}{\textbf{BS=8}} \\
        \cmidrule(lr){3-4} \cmidrule(lr){5-6}
         & & \textbf{TPS} & \textbf{ACC} & \textbf{TPS} & \textbf{ACC} \\
        \midrule
        \textbf{Dream} & 16 & 178.0 & \textbf{77.48} & 204.3 & \textbf{78.62} \\
         & 32 & \textbf{222.9} & 76.50 & \textbf{263.3} & 74.68 \\
         & 64 & 213.1 & 67.48 & 243.1 & 65.50 \\
        \midrule
        \textbf{LLaDA} & 16 & 184.0 & \textbf{79.00} & 232.0 & \textbf{79.08} \\
         & 32 & 248.8 & 77.56 & 330.4 & 77.48 \\
         & 64 & \textbf{280.8} & 77.33 & \textbf{366.1} & 76.95 \\
        \bottomrule
        \end{tabular}
        }
    \end{minipage}
    \hfill
    \begin{minipage}[t]{0.48\linewidth}
        \centering
        \caption{Block length ($L$) ablation on HumanEval (Gen Len 256, Batch Size 4/8).}
        \label{tab:ablation_block_size_humaneval}
        \resizebox{\linewidth}{!}{
        \begin{tabular}{lccccc}
        \toprule
        \multirow{2}{*}{\textbf{Model}} & \multirow{2}{*}{$L$} & \multicolumn{2}{c}{\textbf{BS=4}} & \multicolumn{2}{c}{\textbf{BS=8}} \\
        \cmidrule(lr){3-4} \cmidrule(lr){5-6}
         & & \textbf{TPS} & \textbf{ACC} & \textbf{TPS} & \textbf{ACC} \\
        \midrule
        \textbf{Dream} & 16 & 268.6 & 54.88 & 381.4 & 53.66 \\
         & 32 & \textbf{293.9} & \textbf{57.32} & \textbf{407.3} & \textbf{58.54} \\
         & 64 & 262.1 & 55.49 & 314.2 & 57.32 \\
        \midrule
        \textbf{LLaDA} & 16 & 250.8 & 36.59 & 366.6 & 37.20 \\
         & 32 & \textbf{284.9} & \textbf{38.41} & \textbf{409.8} & \textbf{38.41} \\
         & 64 & 283.5 & 34.15 & 393.0 & 33.54 \\
        \bottomrule
        \end{tabular}
        }
    \end{minipage}
\end{table*}

At $L=16$, throughput is consistently lower than at $L=32$, indicating overhead at fine granularity. Increasing to $L=64$ can degrade accuracy---Dream on GSM8K suffers a $\sim$9\% accuracy drop---because larger blocks denoise more tokens simultaneously, which can amplify the approximation effects inherent in parallel decoding~\cite{wu2025fast}. We adopt $L=32$, balancing scheduling overhead against generation stability.

\paragraph{Impact of prompt-length sorting}
The admission controller sorts the pending queue by prompt length to reduce padding overhead. To evaluate this effect, we compare sorted and unsorted queues with all other parameters held constant.

Table~\ref{tab:ablation_sorting} reports results at BS=16 across both models. Length-aware sorting consistently improves throughput by 4.7\%--9.2\%. This aligns with the analysis in Section~\ref{sec:method:admission}: grouping requests of similar prompt length ensures that newly admitted requests have comparable initial row widths, resulting in a tighter bounding box with less padding waste.

\begin{table*}[t]
\centering
\small
\caption{Prompt-length sorting ablation (BS=16). Sorted: queue ordered by prompt length. Unsorted: unordered. Gen len: 256 (GSM8K, HumanEval), 512 (MATH).}
\label{tab:ablation_sorting}
\resizebox{\textwidth}{!}{
\begin{tabular}{llccccc}
\toprule
\textbf{Model} & \textbf{Dataset} & \textbf{TPS (Unsorted)} & \textbf{TPS (Sorted)} & \textbf{Improvement} & \textbf{Time (Unsorted)} & \textbf{Time (Sorted)} \\
\midrule
\multirow{3}{*}{\textbf{Dream}} & GSM8K & 260.3 & 279.0 & 7.2\% & 669.5 s & 626.2 s \\
 & HumanEval & 424.0 & 462.9 & 9.2\% & 48.8 s & 45.2 s \\
 & MATH & 285.3 & 305.4 & 7.0\% & 3998.2 s & 3725.4 s \\
\midrule
\multirow{3}{*}{\textbf{LLaDA}} & GSM8K & 351.5 & 378.9 & 7.8\% & 852.0 s & 807.0 s \\
 & HumanEval & 519.7 & 552.6 & 6.3\% & 72.0 s & 67.8 s \\
 & MATH & 369.6 & 386.8 & 4.7\% & 5768.5 s & 5547.5 s \\
\bottomrule
\end{tabular}
}
\end{table*}

\section{Related Work}
\label{sec:related_work}

\subsection{Diffusion Language Models}

Diffusion models have emerged as a compelling alternative to autoregressive (AR) language models, producing text by iteratively denoising a corrupted sequence rather than generating tokens sequentially.
Early discrete diffusion foundations---structured transition matrices~\cite{austin2021structured}, score entropy~\cite{lou2023discrete}, and masked diffusion with semi-autoregressive samplers~\cite{sahoo2024simple}---progressively narrowed the quality gap with AR models.
Building on these advances, LLaDA~\cite{nie2025large} scaled masked diffusion to 8B parameters with semi-autoregressive block generation, achieving performance competitive with LLaMA3~8B.
Ye et al.~\cite{ye2024beyond} demonstrated that discrete diffusion models are particularly effective for complex reasoning and planning tasks.
Dream~\cite{ye2025dream} further advanced this line by training a 7B-parameter diffusion LLM with AR-based initialization and context-adaptive noise rescheduling, establishing a new state of the art among open diffusion language models. BlockServe is evaluated on both LLaDA and Dream.

\subsection{LLM Serving Systems}

The AR serving ecosystem has converged on iteration-level scheduling~\cite{yu2022orca}, paged KV cache management~\cite{kwon2023efficient}, chunked prefills with stall-free batching~\cite{agrawal2023sarathi,agrawal2024taming}, and prefill-decode disaggregation~\cite{zhong2024distserve}.
These techniques are tailored to the AR execution model, where each iteration produces exactly one token per request.
In contrast, dLLMs exhibit a fundamentally different execution pattern: each request undergoes multiple denoising iterations over a fixed-length block.
This mismatch renders token-level continuous batching inapplicable to dLLM serving, motivating the block-grained scheduling introduced in BlockServe.

\subsection{Diffusion LLM Inference Optimization}

Fast-dLLM~\cite{wu2025fast} accelerates single-request dLLM inference via a block-wise approximate KV cache enabling computation reuse across denoising steps despite bidirectional attention, and a confidence-aware parallel decoding scheme that selectively unmasks high-confidence tokens to reduce denoising iterations.
Concurrently, dLLM-Serve~\cite{fan2025taming} targets the memory footprint challenge in production dLLM serving, introducing logit-aware activation budgeting to cap peak memory from output projections, a phase-multiplexed scheduler that interleaves compute-bound and bandwidth-bound denoising phases across requests, and head-centric sparse attention.

While Fast-dLLM optimizes single-request latency and dLLM-Serve addresses phase-level resource heterogeneity in continuous dLLM serving, neither explicitly targets the convergence heterogeneity that arises when batched requests complete at different block boundaries.
BlockServe fills this gap with block-grained preemption---immediately evicting completed requests to eliminate the straggler effect---mixed-state memory alignment for concurrent requests at different denoising stages, and compute-aware admission control that dynamically adapts batch concurrency to workload geometry.
The single-request optimizations of Fast-dLLM are complementary to BlockServe and are integrated as drop-in accelerators within our scheduling framework without modification.

\section{Conclusion}
\label{sec:conclusion}

BlockServe demonstrates that block-grained scheduling substantially alleviates the straggler bottleneck in batched diffusion LLM serving, achieving 1.9--10.6$\times$ throughput improvement across two models compared with the best-performing baseline we evaluated, while maintaining comparable generation quality.
The core insight is that convergence heterogeneity---the varying rates at which dLLM requests reach completion---can be addressed by treating each execution block as an independent scheduling quantum.
By combining block-level eviction with mixed-state memory alignment, BlockServe keeps GPU resources better utilized even as individual requests complete at different paces.
Token-budget admission control further enables a 2--4$\times$ increase in effective batch capacity under the same hardware constraints.
Our evaluation focuses on offline batch inference; extending BlockServe to online serving with dynamic request arrivals remains future work.

\bibliographystyle{IEEEtran}
\bibliography{references}

\end{document}